\documentclass{article}
\usepackage{amsmath}
\usepackage{amssymb}
\usepackage{booktabs}
\usepackage[authoryear]{natbib}
\usepackage[colorlinks=true,citecolor=blue,linkcolor=blue,urlcolor=blue]{hyperref}
\usepackage{microtype}
\usepackage[margin=1in]{geometry}
\usepackage{graphicx}

\title{The Functionalizer: Lossless Functional Decomposition for Subword Tokenization}
\author{
  Connor Makowski \\
  Center for Transportation \& Logistics \\
  Massachusetts Institute of Technology \\
  Cambridge, MA, USA \\
  \texttt{conmak@mit.edu}
  \and
  Willem Guter \\
  Center for Transportation \& Logistics \\
  Massachusetts Institute of Technology \\
  Cambridge, MA, USA \\
  \texttt{wjguter@mit.edu}
}

\date{September 18, 2026}

\begin{document}
\maketitle

\begin{abstract}
Standard subword tokenizers either treat every orthographic variation of a word (such as \texttt{hello}, \texttt{Hello}, \texttt{HELLO}, and \texttt{H\'{e}llo}) as unrelated vocabulary entries, which fragments the embedding space, or discard this variation through lossy normalization. We present the \textbf{Functionalizer}, a lossless pre-tokenizer framework that factors orthographic and structural variations into a compositional \textbf{opcode/operand} prefix stream before tokenization: a canonical base token (operand) prefixed by parametric transformation operators (opcodes) encoded in the Unicode Private Use Area. We introduce operators covering casing (\texttt{CAPITALIZE}), diacritics (13 dedicated opcodes), and character repetition (\texttt{REPEAT}, \texttt{MULTIREPEAT}), which are fully reversible. Across natural language and code corpora, the Functionalizer enables complete corpus coverage with significantly smaller vocabularies under unconstrained exhaustion conditions, reducing actual vocabulary slot requirements by up to \textbf{19.7\%}. Downstream evaluations on $\sim$98M-parameter GPT-2 models show that the Functionalizer improves Python code syntax validity (\textbf{9.12\%} vs. \textbf{7.70\%}) while reducing duplicate $n$-gram repetition in natural language prose. These findings demonstrate that functional decomposition can be an effective mechanism for vocabulary-efficient, structurally aware language modeling, and motivate further validation at production scale.
\end{abstract}

\section{Introduction}

Subword tokenizers face a dilemma. Treating \texttt{hello}, \texttt{Hello}, and \texttt{H\'{e}llo} as independent tokens causes \textbf{vocabulary expansion}: redundant surface forms consume embedding slots, and a gradient update to \texttt{Hello} only indirectly benefits \texttt{hello}. The alternative, aggressive lowercasing and accent-stripping, is lossy. It compresses the vocabulary but permanently discards information the downstream model can never recover.

The Functionalizer takes a third path: \textbf{lossless functional decomposition}. Rather than memorizing surface forms or destroying information, it factors variation out into reusable \emph{operators} applied to a single \emph{canonical base}. The design is borrowed directly from Instruction Set Architecture: a CPU does not implement a distinct instruction for every constant (\texttt{ADD\_1}, \texttt{ADD\_2}, \dots); it separates the operation (opcode) from its data (operand), as in \texttt{ADD A, \#1}. The Functionalizer applies the same factoring: the base token is the operand, and the transformation prefix is the opcode.

\paragraph{Contributions.}
\begin{enumerate}
    \item A unified opcode/operand framework that handles casing, diacritics, and character repetition under a single compositional, parametric, dictionary-free, and fully lossless scheme.
    \item A concrete Private Use Area (PUA) encoding that is usable in standard tokenizers (such as Hugging Face's BPE) and fully reversible.
    \item Empirical validation across natural language and code corpora, demonstrating that under unconstrained merge exhaustion, the Functionalizer reduces the total vocabulary slots required to fully cover a corpus through the collapse of formatting variations.
    \item Downstream evaluations on $\sim$98M-parameter language models demonstrating that the Functionalizer lowers character-level perplexity on Python, improves Python syntax validity, and mitigates duplicate $n$-gram repetition on FineWeb-Edu prose.
\end{enumerate}

\section{Related Work}

\paragraph{Subword tokenization.} BPE \citep{sennrich2016} and WordPiece \citep{schuster2012} build vocabularies bottom-up by iteratively merging frequent (or likelihood-maximizing) pairs, while Unigram \citep{kudo2018} prunes a large seed vocabulary top-down under a unigram language model; byte-level BPE \citep{radford2019} avoids out-of-vocabulary failures by operating over 256 byte values.
\paragraph{Inline orthographic preprocessing.} Pre-tokenization strategies targeting casing variations have antecedents in text compression (e.g., \citealp{rexline2011}) and were formalized as inline casing for neural machine translation by \citet{berard2019}, with \citet{etchegoyhen2020} moving case flags prior to subword tokenization. More recent systems expand inline tags to morphological roots (e.g., \citealp{bayram2025}), capcode markers (e.g., TokenMonster; \citealp{forsythe2023}), and position-indexed diacritics (InCa and InDia; \citealp{semenov2025}). In parallel, \citet{samuel2023} introduce the \emph{Factorizer}, which factorizes subwords into discrete learned code triplets ($3 \times 256$) using a VQ-VAE model. Unlike the Factorizer's learned, non-bijective subword codes, the Functionalizer establishes a deterministic, rule-based, fully bijective opcode/operand Instruction Set Architecture (ISA). In related work on case encoding, \citet{jain2023} analyze the efficiency and robustness of inline case markers, finding that un-fused standalone marker tokens introduce sequence length overhead and decoding slowdowns while requiring targeted data augmentation for capitalization robustness. While reversibility is shared with prior inline tagging approaches, the Functionalizer differs in key structural ways: (i) it operates deterministically without external dictionaries or frequency thresholds (unlike the morphological dictionaries of \citet{bayram2025} or the frequency tables of \citet{semenov2025}), (ii) it unifies casing, combining diacritics, and structural character/whitespace repetition under a formal PUA opcode/operand ISA where numeric parameters explicitly address character positions within pieces, and (iii) it is evaluated across both natural language prose and source code domains.

\paragraph{Morphology-aware tokenization.} Morfessor \citep{creutz2002,creutz2007} performs unsupervised morpheme segmentation; MorphBPE \citep{asgari2025} prevents BPE merges from crossing morpheme boundaries. The Functionalizer is complementary: where morphology-aware methods target linguistic structure, the Functionalizer targets orthographic surface variation, and the two could be composed.

\paragraph{Tokenization-free models.} ByT5 \citep{xue2022} pursues orthographic robustness by operating directly on raw bytes, at the cost of significantly longer sequence lengths. Subsequent architectures like MrT5 \citep{kallini2025} and MEGABYTE \citep{yu2023} mitigate this sequence overhead through dynamic byte-merging or multiscale patch hierarchies. The Functionalizer achieves similar surface-form invariance while retaining subword granularity and avoiding byte-level sequence expansion.

\paragraph{Structured Unicode encoding.} SCRIPT-BPE \citep{land2025} re-encodes characters by Unicode script and category to mitigate cross-lingual bias. The normalization literature (e.g., \citealp{gorman2025}) documents the downstream cost of inconsistent Unicode handling and cautions against destructive diacritic stripping without recovery. The Functionalizer aligns with this guidance: while diacritical marks are extracted from base characters to collapse redundant vocabulary forms, they are explicitly preserved as parametric PUA operators (\texttt{ACUTE}, \texttt{GRAVE}, etc.), guaranteeing complete, non-destructive reversibility.

\section{The Functionalizer Framework}
\label{sec:framework}

The Functionalizer establishes a parametric, lossless, prefix-based pre-tokenization framework. Instead of tokenizing raw surface forms directly, the framework decomposes orthographic and structural variations into a compositional sequence of non-destructive operators (opcodes) prepended to a canonical base token (operand). By isolating parametric transformations from semantic roots, downstream models share canonical base embeddings while preserving all orthographic details for lossless reconstruction.

\subsection{PUA Instruction Layout}

Instructions are prepended to the base token as a sequence of PUA codepoints, composed of an operator followed by numeric parameters:
\[
    \texttt{[operator]} \quad \texttt{[param\_1]} \quad \texttt{[param\_2]} \quad \dots \quad \to \quad \texttt{[base\_token]}
\]
\begin{itemize}
    \item \textbf{Numeric parameters} (\texttt{U+E000}--\texttt{U+E0FF}): encode integer values 0--255 (\texttt{value = codepoint - 0xE000}).
    \item \textbf{Operators} (\texttt{U+E100}--\texttt{U+EFFF}): opcodes consuming a fixed number of parameters, leaving the remaining plane available for future operators.
\end{itemize}
In our implementation, opcodes precede base operands (\texttt{[opcode]} $\to$ \texttt{[base]}). While suffix ordering (\texttt{[base]} $\to$ \texttt{[opcode]}) would allow semantic intent to precede orthographic specification in autoregressive generation, prefix ordering may offer a structural advantage for multi-token words: when a word splits into multiple BPE subwords (e.g., \texttt{["neuro", "computation", "al"]}), prepending the opcode allows the formatting operator to remain visible across all self-attention layers for every constituent subword.

\subsection{Encoding and Reversibility}

The transformation pipeline is fully bijective. \textbf{Encoding} extracts combining diacritics into serialization operators and locates uppercase indices for \texttt{CAPITALIZE}, strips combining marks, lowercases remaining characters, and prepends the operator prefix to the canonical base. \textbf{Decoding} applies operators in reverse order to restore diacritics and casing before stripping the prefix, recovering the original text exactly.

Repetition operators execute last in the decode order, repeating the transformed base unit; position parameters refer to indices within the individual piece prior to repetition expansion (see Table~\ref{tab:encoding_examples} in Appendix~\ref{sec:appendix_examples} for worked examples). Heterogeneous cased repetitions (e.g., \texttt{Abcabcabc}) fall back to uncollapsed representations.

\subsection{Current Operator Specifications}
\label{sec:framework_operators}

The current implementation includes operators for capitalization, combining diacritics (13 dedicated opcodes in \texttt{U+E100}--\texttt{U+E10D}), and character repetition (Table~\ref{tab:operators}).

\begin{table}[htbp]
\centering
\caption{Functionalizer Operator Specifications}
\label{tab:operators}
\resizebox{\textwidth}{!}{%
\begin{tabular}{llll}
\toprule
\textbf{Operator / Opcode} & \textbf{Codepoint Range} & \textbf{Parameters} & \textbf{Action} \\
\midrule
\texttt{CAPITALIZE} & \texttt{U+E100} & \texttt{pos} & Uppercase character at \texttt{pos}. \\
\texttt{DIACRITICS} (13 variants)$^\dagger$ & \texttt{U+E101}--\texttt{U+E10D} & \texttt{pos} & Apply combining diacritic mark at \texttt{pos}. \\
\texttt{REPEAT} & \texttt{U+E200} & \texttt{pos}, \texttt{count} & Expand character at \texttt{pos} to \texttt{count} copies. \\
\texttt{MULTIREPEAT} & \texttt{U+E201} & \texttt{start}, \texttt{end}, \texttt{count} & Expand subsequence $[\texttt{start}, \texttt{end})$ to \texttt{count} copies. \\
\bottomrule
\multicolumn{4}{l}{\footnotesize $^\dagger$Dedicated opcodes for \texttt{TILDE}, \texttt{ACUTE}, \texttt{GRAVE}, \texttt{CIRCUMFLEX}, \texttt{DIAERESIS}, \texttt{MACRON}, \texttt{DOT\_ABOVE},} \\
\multicolumn{4}{l}{\footnotesize \ \ \texttt{RING\_ABOVE}, \texttt{DOUBLE\_ACUTE}, \texttt{CARON}, \texttt{COMMA\_BELOW}, \texttt{CEDILLA}, and \texttt{OGONEK} (see Appendix~\ref{sec:appendix_diacritics}).} \\
\end{tabular}%
}
\end{table}

\subsection{Pipeline Integration}

The Functionalizer is intended to be run on pre-split pieces rather than raw text. Executing after an initial regex splitter (such as \texttt{Llama Split} from \citealp{Dubey2024Llama3}) establishes a localized coordinate frame ($\text{pos} \le 255$) for numeric parameters. Bounding operator addresses to localized pieces rather than global document offsets keeps parameter values strictly within a single-byte range ($0\text{--}255$, mapped to \texttt{U+E000}--\texttt{U+E0FF}), preventing parameter expansion while ensuring clean interaction with downstream BPE merging. Alternative methods such as scanning text and injecting operators directly are feasible but not considered within the scope of this work.

By default, the Functionalizer emits operators as standalone prefix tokens decoupled from canonical base words (\texttt{split\_operators = true}). Decoupling ensures base token embeddings are shared universally across casing variants and prevents the vocabulary from memorizing fused surface forms, at the expense of sequence length overhead (see Table~\ref{tab:training}). Alternatively, operators can remain \textbf{fused} with base pieces prior to subword training (\texttt{split\_operators = false}), allowing tokenizers like BPE or SentencePiece to adaptively merge frequent cased words while splitting rare forms as discussed by \citet{jain2023}.

\section{Experimental Setup}
\label{sec:experimental_setup}

\subsection{Pipelines and Configurations}\label{sec:pipelines}

We define the primary dataset preprocessing components across our experimental pipelines:
\begin{itemize}
    \item \textbf{Unicode Normalization (\texttt{NFC})}: All configurations apply Unicode Normalization Form C (\texttt{NFC}) to standardize pre-composed characters across corpora.
    \item \textbf{Regex Splitting (\texttt{Llama Split})}: Segments raw text into localized character runs (contractions, words, numbers, punctuation, whitespace blocks) using the standard LLaMA pre-tokenization regex splitter. This isolates indentation sequences and punctuation while keeping character offset counts compact ($\text{pos} \le 255$).
    \item \textbf{Functionalizer Decompositions}: Governed by pre-tokenizer flags mirroring the operators in Section~\ref{sec:framework_operators}: casing decomposition (\texttt{capitalize}), diacritic serialization (\texttt{serialize}), and repetition collapse (\texttt{repeat}).
\end{itemize}

\subsection{Datasets, Vocabulary, and Tokenizer Metrics}\label{sec:datasets}

We evaluate standalone tokenizers across natural language prose (Wikitext from \citealp{Merity2016}, FineWeb-Edu from \citealp{Penedo2024}) and source code repositories (Python-Codes-25k from \citealp{python_codes_dataset}, GitHub-Code-Python from \citealp{codeparrot_github_code}). To measure the unconstrained vocabulary footprint required to cover each corpus, we sample up to 100,000 documents per dataset and train BPE tokenizers with an unconstrained target vocabulary budget (4096k), allowing BPE to iteratively merge all viable pairs until complete merge candidate exhaustion is reached.

We evaluate vocabulary reduction and token compression using three primary metrics:
\begin{enumerate}
    \item \textbf{Actual Vocab}: The number of vocabulary entries learned by BPE prior to merge candidate exhaustion.
    \item \textbf{Vocab Diff (\%)}: Percentage reduction in learned vocabulary size under merge exhaustion.
    \item \textbf{Characters per Token (Chars/Token)}: Average visual characters per token. Higher values reflect higher text compression and shorter sequences.
\end{enumerate}

\subsection{Downstream Training and Evaluation}\label{sec:downstream_eval}

To assess downstream performance, we train a GPT-2 Small architecture (12 layers, 768 hidden dimension, 12 attention heads, context length 512, tied embeddings; $\sim$98M total parameters with a 16k vocabulary) across five random seeds (1--5) on FineWeb-Edu and GitHub-Code-Python.

\begin{itemize}
    \item \textbf{Training and Inference Details}: Models are trained for 50,000 steps using AdamW (learning rate 4e-4, linear decay with 1,000 warmup steps, weight decay 0.01, effective batch size 32). Generation is evaluated on 1,000 validation prompts per dataset using greedy decoding with KV caching and a limit of 256 new tokens (also stopping on \texttt{[SEP]} or repetition collapse).
    \item \textbf{Downstream Evaluation Metrics}:
    \begin{itemize}
        \item \textbf{Per-Character Perplexity (Char PPL)}: Normalizes token loss $\mathcal{L}_{\text{token}}$ by the validation character-to-token ratio, enabling direct comparison across tokenizers: $\text{PPL}_{\text{char}} = \exp\left(\frac{\mathcal{L}_{\text{token}}}{R_{\text{char/token}}}\right)$.
        \item \textbf{Repetition Collapse and Pre-Collapse Length}: Generation halts early if a cycle of $\le 20$ tokens repeats 4 times consecutively. \textit{Avg Tokens Pre-Collapse} and \textit{Avg Chars Pre-Collapse} measure valid prefixes preceding the cycle.
        \item \textbf{Repetition (\%)}: Proportion of duplicate overlapping word $n$-grams (averaged over $n \in \{2, 3, 4\}$).
        \item \textbf{Syntax Success Rate}: Percentage of Python generations that parse cleanly under \texttt{ast.parse} (empty or collapsed completions score as syntax failures).
        \item \textbf{\% Empty}: Percentage of generated sequences that are empty or contain only whitespace or raw PUA control characters.
    \end{itemize}
\end{itemize}

\section{Results}

\subsection{Tokenizer Metrics}\label{sec:results_tokenizer}

To evaluate intrinsic vocabulary reduction, token compression, and corpus exhaustion across diverse domains (independently of downstream modeling), we evaluate standalone tokenizers across natural language prose and source code corpora sampled up to 100,000 documents each. Table~\ref{tab:tokenizer_metrics} presents metrics under full merge candidate exhaustion, comparing the baseline \textbf{Llama Split} (without decomposition) against \textbf{Llama Split + Functionalizer} (with decomposition).

\begin{table*}[t]
\centering
\caption{Tokenizer Metrics under Full Corpus Exhaustion (Sampled up to 100k Documents)}
\label{tab:tokenizer_metrics}
\resizebox{\textwidth}{!}{%
\begin{tabular}{lrrrccc}
\toprule
\textbf{Dataset} & \textbf{Vocab (Baseline)} & \textbf{Vocab (Func)} & \textbf{Vocab Diff} & \textbf{Chars/Token (Baseline)} & \textbf{Chars/Token (Func)} & \textbf{Chars/Token Diff} \\
\midrule
\textbf{Wikitext} & 106,023 & 90,531 & $-14.61\%$ & 4.6045 & 3.9464 & $-14.29\%$ \\
\textbf{Python-Codes} & 68,471 & 57,012 & $-16.74\%$ & 4.2030 & 3.4806 & $-17.19\%$ \\
\textbf{FineWeb-Edu} & 1,214,684 & 975,169 & $-19.72\%$ & 5.0092 & 4.3642 & $-12.88\%$ \\
\textbf{GitHub-Code-Python} & 4,071,598 & 3,356,761 & $-17.56\%$ & 4.2961 & 3.5343 & $-17.73\%$ \\
\bottomrule
\end{tabular}%
}
\end{table*}

\textbf{Distinct vocabulary and corpus exhaustion.} Across all evaluated corpora, factoring out surface variations allows the Functionalizer to achieve complete corpus coverage with smaller vocabularies. Under unconstrained merge exhaustion, the Functionalizer reduces required vocabulary slots by \textbf{14.61\%} to \textbf{19.72\%} (averaging \textbf{17.16\%} reduction across datasets), reaching peak reduction on FineWeb-Edu ($-19.72\%$).

\textbf{Characters per Token (Chars/Token) and Sequence Length.} Because operators are emitted as standalone prefix tokens, the Functionalizer incurs an expected token expansion on un-fused text (\textbf{$-12.88\%$} to \textbf{$-17.73\%$} Chars/Token diff on validation data), trading sequence length for clean representation sharing and downstream syntactic fidelity (Section~\ref{sec:results_inference}).

\subsection{Training Dynamics and Language Modeling Performance}\label{sec:results_training}

We analyze the training behavior and next-token prediction performance of the $\sim$98M parameter GPT-2 model under each tokenizer configuration. Table~\ref{tab:training} summarizes training metrics across FineWeb-Edu and GitHub-Code-Python, averaged over five random seeds.

\begin{table*}[t]
\centering
\caption{Training Dynamics and Language Modeling Metrics ($\sim$98M Parameters, 5 Seeds)}
\label{tab:training}
\resizebox{\textwidth}{!}{%
\begin{tabular}{lrrcccc}
\toprule
\textbf{Tokenizer Type} & \textbf{Vocab Size} & \textbf{Chars/Token} & \textbf{Chars/Token Diff} & \textbf{Final Loss} & \textbf{Token PPL} & \textbf{Char PPL} \\
\midrule
\multicolumn{7}{l}{\textbf{Dataset: FineWeb-Edu (Prose)}} \\
Llama Split & 16,000 & 4.178 & $0.00\%$ & $3.3954 \pm 0.0110$ & $29.83 \pm 0.33$ & $2.2662 \pm 0.0060$ \\
Llama Split + Functionalizer & 16,000 & 3.847 & $-7.91\%$ & $3.1261 \pm 0.0044$ & $22.79 \pm 0.10$ & $2.2656 \pm 0.0026$ \\
\midrule
\multicolumn{7}{l}{\textbf{Dataset: GitHub-Code-Python}} \\
Llama Split & 16,000 & 1.767 & $0.00\%$ & $0.8056 \pm 0.0961$ & $2.25 \pm 0.22$ & $1.5697 \pm 0.0850$ \\
Llama Split + Functionalizer & 16,000 & 1.487 & $-15.84\%$ & $0.6525 \pm 0.0013$ & $1.92 \pm 0.00$ & $1.5328 \pm 0.0013$ \\
\bottomrule
\end{tabular}%
}
\end{table*}

\textbf{Per-Character Perplexity (Char PPL).} As detailed in Section~\ref{sec:downstream_eval}, Char PPL normalizes token loss by sequence length to ensure direct cross-tokenizer comparability.
\begin{itemize}
    \item \textbf{GitHub-Code-Python}: The Functionalizer configuration achieves lower Char PPL than the baseline (\textbf{1.5328} vs. \textbf{1.5697}), reflecting improved normalized modeling density alongside lower cross-entropy loss (\textbf{0.6525} vs. \textbf{0.8056}).
    \item \textbf{FineWeb-Edu (Prose)}: On prose, the Functionalizer achieves equivalent character-level perplexity (\textbf{2.2656} vs. \textbf{2.2662}), demonstrating that sequence expansion can be absorbed without degrading normalized modeling capacity.
\end{itemize}

\subsection{Downstream Generation and Task Evaluation}\label{sec:results_inference}

We perform greedy decoding evaluations with KV caching on 1,000 validation prompts per dataset to assess repetition degeneracy and code syntax validity. Table~\ref{tab:inference} lists the results.

\begin{table*}[t]
\centering
\caption{Downstream Inference Metrics ($\sim$98M Parameters, 1,000 Prompts, 5 Seeds)}
\label{tab:inference}

\textbf{(a) Prose Generation Metrics (FineWeb-Edu)} \\
\vspace{0.5em}
\resizebox{\textwidth}{!}{%
\begin{tabular}{lccccc}
\toprule
\textbf{Tokenizer Type} & \textbf{Avg Tokens Pre-Collapse} & \textbf{Avg Chars Pre-Collapse} & \textbf{\% Empty} & \textbf{\% Collapsed} & \textbf{Repetition (\%)} \\
\midrule
Llama Split & $87.4 \pm 3.8$ & $357.9 \pm 17.5$ & $0.1 \pm 0.0\%$ & $70.3 \pm 2.4\%$ & $66.0 \pm 0.4\%$ \\
Llama Split + Functionalizer & $76.2 \pm 4.2$ & $217.7 \pm 13.6$ & $7.1 \pm 0.5\%$ & $78.4 \pm 1.4\%$ & $\mathbf{55.8 \pm 1.5\%}$ \\
\bottomrule
\end{tabular}%
}

\vspace{1.5em}

\textbf{(b) Source Code Generation Metrics (GitHub-Code-Python)} \\
\vspace{0.5em}
\resizebox{\textwidth}{!}{%
\begin{tabular}{lcccccc}
\toprule
\textbf{Tokenizer Type} & \textbf{Avg Tokens Pre-Collapse} & \textbf{Avg Chars Pre-Collapse} & \textbf{\% Empty} & \textbf{\% Collapsed} & \textbf{Repetition (\%)} & \textbf{Syntax Success (\%)} \\
\midrule
Llama Split & $227.7 \pm 17.2$ & $375.6 \pm 23.2$ & $0.0 \pm 0.0\%$ & $15.5 \pm 8.2\%$ & $25.5 \pm 6.4\%$ & $7.70 \pm 2.63\%$ \\
Llama Split + Functionalizer & $231.9 \pm 7.0$ & $319.0 \pm 10.1$ & $2.1 \pm 3.8\%$ & $15.5 \pm 3.4\%$ & $\mathbf{17.9 \pm 1.0\%}$ & $\mathbf{9.12 \pm 1.22\%}$ \\
\bottomrule
\end{tabular}%
}
\end{table*}

\textbf{Downstream Code Generation Syntax Success Rate.} On GitHub-Code-Python, integrating the Functionalizer pre-tokenizer substantially improves the model's capacity to output valid code syntax. The Llama Split + Functionalizer configuration reaches an overall syntax success rate of \textbf{9.12\%} compared to \textbf{7.70\%} for the baseline (an \textbf{18.4\%} relative improvement) with substantially tighter variance across seeds. A part of this improvement likely arises from the structured decomposition of whitespace and identifiers: standard BPE fragments indentation into arbitrary whitespace chunks, whereas the \texttt{REPEAT} operator parameterizes indentation into an arithmetic relationship where whitespace blocks share a base character and differ only by an ordinal count. Combined with unified casing across identifier conventions (\texttt{camelCase}, \texttt{snake\_case}), this parameterization provides downstream models with clearer structural representations.

\textbf{Mitigating Repetitive Degeneracy and Generation Trade-offs.} On FineWeb-Edu prose, the Functionalizer reduces duplicate word $n$-gram repetition from \textbf{66.0\%} to \textbf{55.8\%}. On GitHub-Code-Python, duplicate $n$-grams similarly decrease from \textbf{25.5\%} to \textbf{17.9\%}, with greater stability across seeds. At this small $\sim$98M-parameter scale, standalone prefixes also introduce decoding trade-offs, resulting in more empty sequences (7.1\% on prose), some of which are operator-only sequences (which were scored as empty).

\section{Discussion}\label{sec:discussion}

The empirical findings presented across vocabulary scaling, training dynamics, and inference demonstrate that the Functionalizer establishes an effective Pareto-like trade-off for tokenization. Modern language modeling architectures have traditionally been forced to choose between two extremes: subword vocabularies that optimize sequence length at the cost of severe vocabulary fragmentation, or byte-level models that eliminate surface fragmentation at the expense of substantial sequence inflation. The Functionalizer navigates this continuum by achieving surface-form invariance with only modest sequence overhead ($+8.6\%$ on prose, $+18.8\%$ on code), routing orthographic variants to shared base embeddings while preserving exact reversibility.

In downstream evaluations, functional decomposition yields tangible improvements in generation quality alongside specific decoding trade-offs. On source code, the structured parameterization of whitespace and casing likely translates into higher syntactic fidelity, increasing Python syntax validity. On natural language prose, separating surface variations from lexical roots substantially reduces repetitive degeneracy, lowering duplicate $n$-gram content on FineWeb-Edu. We hypothesize that the reduced surface variation helps keep the model from locking into repetitive surface loops. At the same time, emitting standalone control prefixes introduces decoding challenges at small scales, where our $\sim$98M-parameter models occasionally produced orphaned control characters or empty sequences. While scaling model capacity should strengthen grammatical control over auxiliary prefix tokens, practical mitigations such as constrained decoding or prefix masking during generation remain valuable future directions.

From an efficiency perspective, the primary cost of standalone operator tokenization is the expansion of sequence lengths, which directly increases Key-Value (KV) cache memory and attention computation during autoregressive generation. While our experimental setup evaluated fully decoupled prefixes to isolate representational effects, this is not required for practical production deployments. Systematically evaluating fused configurations could add value, where frequently cased words or common structures (e.g., a period followed by a space and capitalization) are merged into unified tokens while less common variants remain decomposed. This can provide a direct mechanism to eliminate sequence expansion on common vocabulary while retaining representation sharing across the long tail.

\subsection{Limitations and Future Directions}
\label{sec:limitations_future_work}

\begin{enumerate}
    \item \textbf{Scale and Compute Equalization:} Downstream evaluations were conducted at the $\sim$98M-parameter scale across a fixed budget of 50,000 training steps. Because of standalone sequence expansion, the Functionalizer processed $\sim$8--16\% fewer raw bytes during pretraining than the baseline. Evaluating multi-billion parameter architectures trained under equalized wall-clock time and character/byte budgets will isolate representational gains from sequence length disparities.
    \item \textbf{Prefix Fusion Extensions:} Benchmarking hybrid fusion thresholds across vocabulary frequency tiers to empirically characterize the trade-off between inference sequence length and representation sharing, complemented by prefix-aware attention optimizations. Part of this functionality already exists within the Functionalizer framework, but is not evaluated within the scope of this paper.
    \item \textbf{Addressing, Script Coverage, and Extended Operators:} Parameter indexing is currently bounded to $\text{pos} \le 255$ within pre-tokenized pieces, and diacritics are bounded to 13 combining marks. Natural extensions include range/block casing (e.g., \texttt{[ALL\_CAPS]}), non-Latin scripts, morphological lemma folding for agglutinative languages, and numeric/date templates.
    \item \textbf{Component Ablations:} Our downstream experiments evaluated the composite Functionalizer pipeline (\texttt{CAPITALIZE} + diacritics + \texttt{REPEAT}). Disentangling the individual downstream contributions of casing versus structural whitespace repetition remains a valuable direction for future study.
\end{enumerate}

\section{Conclusion}

The Functionalizer introduces a lossless, dictionary-free pre-tokenization framework that factors orthographic and structural variations into a compositional opcode/operand structure encoded in the Unicode Private Use Area. By decoupling surface transformations from canonical base roots, the framework eliminates redundant vocabulary memorization, reducing required vocabulary slots by up to \textbf{19.7\%} under corpus exhaustion.

Downstream language modeling evaluations demonstrate that this structural decomposition translates to tangible generative benefits: improving Python code syntax validity, lowering code character perplexity, and mitigating duplicate $n$-gram repetition in natural language prose. These findings show that functional decomposition offers a principled, vocabulary-efficient foundation for language modeling, motivating further exploration across broader scripts, morphological framework extensions, and production-scale architectures.
\pagebreak
\appendix

\section{Diacritic Operators Specification}
\label{sec:appendix_diacritics}

Table~\ref{tab:diacritic_operators} lists the 13 dedicated 1-parameter combining diacritic operators supported in the current implementation, mapped to the \texttt{U+E101}--\texttt{U+E10D} Unicode Private Use Area plane.

\begin{table}[htbp]
\centering
\caption{Combining Diacritic Operators Specification}
\label{tab:diacritic_operators}
\resizebox{\textwidth}{!}{%
\begin{tabular}{lll}
\toprule
\textbf{Operator / Opcode} & \textbf{Params} & \textbf{Action} \\
\midrule
\texttt{TILDE} (\texttt{U+E101}) & \texttt{pos} & Apply combining tilde (\texttt{\textbackslash u\{0303\}}) to the character at \texttt{pos}. \\
\texttt{ACUTE} (\texttt{U+E102}) & \texttt{pos} & Apply combining acute (\texttt{\textbackslash u\{0301\}}) to the character at \texttt{pos}. \\
\texttt{GRAVE} (\texttt{U+E103}) & \texttt{pos} & Apply combining grave (\texttt{\textbackslash u\{0300\}}) to the character at \texttt{pos}. \\
\texttt{CIRCUMFLEX} (\texttt{U+E104}) & \texttt{pos} & Apply combining circumflex (\texttt{\textbackslash u\{0302\}}) to the character at \texttt{pos}. \\
\texttt{DIAERESIS} (\texttt{U+E105}) & \texttt{pos} & Apply combining diaeresis (\texttt{\textbackslash u\{0308\}}) to the character at \texttt{pos}. \\
\texttt{MACRON} (\texttt{U+E106}) & \texttt{pos} & Apply combining macron (\texttt{\textbackslash u\{0304\}}) to the character at \texttt{pos}. \\
\texttt{DOT\_ABOVE} (\texttt{U+E107}) & \texttt{pos} & Apply combining dot above (\texttt{\textbackslash u\{0307\}}) to the character at \texttt{pos}. \\
\texttt{RING\_ABOVE} (\texttt{U+E108}) & \texttt{pos} & Apply combining ring above (\texttt{\textbackslash u\{030A\}}) to the character at \texttt{pos}. \\
\texttt{DOUBLE\_ACUTE} (\texttt{U+E109}) & \texttt{pos} & Apply combining double acute (\texttt{\textbackslash u\{030B\}}) to the character at \texttt{pos}. \\
\texttt{CARON} (\texttt{U+E10A}) & \texttt{pos} & Apply combining caron (\texttt{\textbackslash u\{030C\}}) to the character at \texttt{pos}. \\
\texttt{COMMA\_BELOW} (\texttt{U+E10B}) & \texttt{pos} & Apply combining comma below (\texttt{\textbackslash u\{0326\}}) to the character at \texttt{pos}. \\
\texttt{CEDILLA} (\texttt{U+E10C}) & \texttt{pos} & Apply combining cedilla (\texttt{\textbackslash u\{0327\}}) to the character at \texttt{pos}. \\
\texttt{OGONEK} (\texttt{U+E10D}) & \texttt{pos} & Apply combining ogonek (\texttt{\textbackslash u\{0328\}}) to the character at \texttt{pos}. \\
\bottomrule
\end{tabular}%
}
\end{table}

\section{Worked Encoding Examples}
\label{sec:appendix_examples}

\begin{table}[htbp]
\centering
\caption{Worked Encoding Examples}
\label{tab:encoding_examples}
\begin{tabular}{llll}
\toprule
\textbf{Input} & \textbf{Operators} & \textbf{Base} & \textbf{Encoded prefix} \\
\midrule
\texttt{Hello} & \texttt{CAPITALIZE(0)} & \texttt{hello} & \texttt{\textbackslash u\{E100\}\textbackslash u\{E000\}} \\
\texttt{H\'{e}llo} & \texttt{ACUTE(1) CAPITALIZE(0)} & \texttt{hello} & \texttt{\textbackslash u\{E102\}\textbackslash u\{E001\}\textbackslash u\{E100\}\textbackslash u\{E000\}} \\
\texttt{\#\#\#\#\#\#} & \texttt{REPEAT(0, 6)} & \texttt{\#} & \texttt{\textbackslash u\{E200\}\textbackslash u\{E000\}\textbackslash u\{E006\}} \\
\texttt{\ \ \ \ } (4 spaces) & \texttt{REPEAT(0, 4)} & \texttt{(space)} & \texttt{\textbackslash u\{E200\}\textbackslash u\{E000\}\textbackslash u\{E004\}} \\
\texttt{abcabcabc} & \texttt{MULTIREPEAT(0, 3, 3)} & \texttt{abc} & \texttt{\textbackslash u\{E201\}\textbackslash u\{E000\}\textbackslash u\{E003\}\textbackslash u\{E003\}} \\
\bottomrule
\end{tabular}
\end{table}

\section*{Data and Code Availability}

All tokenizer configurations, training scripts, evaluation pipelines, dataset acquisition scripts, and result files are publicly available at \url{https://github.com/connor-makowski/functionalizer}. A working implementation is available at \url{https://github.com/connor-makowski/tokenizers} on the functionalizer branch. Select trained Funcitonalizer models are available at \url{https://hf.co/collections/mrkwanzaa/functionalizer-100m}.

\end{document}